# A Deep-Learning Based Optimization Approach to Address Stop-Skipping Strategy in Urban Rail Transit Lines

Mohammadjavad Javadinasr[1], Amir Bahador Parsa[1], and Abolfazl (Kouros) Mohammadian[1]

1) Department of Civil and Materials Engineering, University of Illinois at Chicago, 842 W. Taylor St, Chicago, IL 60607

## ABSTRACT

Different passenger demand rates in transit stations underscore the importance of adopting operational strategies to provide a demand-responsive service. Aiming at improving passengers' travel time, the present study introduces an advanced data-driven optimization approach to determine the optimal stop-skip pattern in urban rail transit lines. In detail, first, using the time-series smart card data for an entire month, we employ a Long Short-Term Memory (LSTM) deep learning model to predict the station-level demand rates for the peak hour. This prediction is based on four preceding hours and is especially important knowing that the true demand rates of the peak hour are posterior information that can be obtained only after the peak hour operation is finished. Moreover, utilizing a real-time prediction instead of assuming fixed demand rates, allows us to account for unexpected real-time changes which can be detrimental to the subsequent analyses. Then, we integrate the output of the LSTM model as an input to an optimization model with the objective of minimizing patrons' total travel time. Considering the exponential nature of the problem, we propose an Ant Colony Optimization technique to solve the problem in a desirable amount of time. Finally, the performance of the proposed models and the solution algorithm is assessed using real case data. The results suggest that the proposed approach can enhance the performance of the service by improving both passengers' in-vehicle time as well as passengers' waiting time.

# 1. Introduction

Transportation systems, are described as the "lifeblood" of cities (Vuchic, 1999), and among which the public transportation systems play a vital role in the economic and social development of every society (Miller et al., 2016). The ability to transport a massive number of people along with affordable prices has made the public transportation systems the main alternative for millions of people. Moreover, the growing population in many metropolitan areas has dramatically increased the demand for transit systems especially urban transit lines (Y. Wang et al., 2014). The headways between consecutive trains at the majority of metro stations in big cities such as Tokyo, New York, Paris, and Beijing are between 3-5 minutes. Aiming at establishing sustainable transit systems (Pezeshknejad et al., 2020), researchers and practitioners seek methods to increase the transit passengers' quality of experience (Soltanpour et al., 2020), which in general term, can be achieved by adopting two types of strategies (Asgharzadeh and Shafahi, 2017):

- Planning strategies.
- Operational control strategies.

Planning strategies refer to structural modifications to improve the transit service (Zhang et al., 2021). Encompassing a wide variety of measures from increasing the fleet size to constructing new routes, the planning strategies require overcoming fiscal and physical constraints (Zolfaghari et al., 2004). On the other hand, focusing on more effective utilization of the existing transit system, operational control strategies aim at slight modifications in the movement of transit vehicles. A large body of literature has explored a variety of operational controls such as vehicle-holding (Dai et al., 2019), deadheading (Wagenaar et al., 2017), zoning (Larrain et al., 2015), short turning (Ding et al., 2018), and stop-skipping (J. Chen et al., 2015). Not only do operational strategies require lower capital investment, but they also are time-efficient alternatives that make them viable solutions to improve the level of service. Although holding strategy is the most studied tactic in the literature (Muñoz et al., 2013), due to its physical requirements, is not easily applicable in many urban transit lines. On the other hand, the stop-skipping strategy has the potential to be employed by both train and bus agencies as a potential contributor to increasing the overall transit service efficiency.

Different stop-to-stop demand rates in urban transit stations underscore the importance of adopting operational strategies to provide a demand-responsive service. Even though boarding new passengers in some stations is hampered by capacity limitation, transit vehicles, specifically subway trains, are traditionally arranged to stop at every station causing longer travel times. Frequent stops along the path and delays generated by deceleration, alighting, boarding, and acceleration can degrade the quality of the service. This is important, especially for the peak hour periods when the oversaturated condition prevents some passengers from boarding, making them wait for the subsequent trains. As a result of this phenomenon, users in overcrowded stations cannot compete with the other passengers boarding at earlier stations for the limited capacity of trains (Shang et al., 2018). In this regard, skipping some stations potentially can benefit both the patrons, by shorter total travel times, increased average speed, and less accumulation of delayed passengers (Zhang et al., 2021), as well as the operators, by lower vehicle depreciation and energy consumption, resulting from higher productivity (Fan and Ran, 2021). The positive effects of the stop-skipping strategy are contingent upon the low deviation between the expected conditions and the daily operation (Gkiotsalitis, 2019). Thus, stop-skipping schemes should be implemented carefully to avoid the potential undesirable repercussions (e.g., increasing travel time for some passengers, bunching, etc.). In this regard, the automated data collection systems, such as automated fare-collection (AFC), in transit networks have provided researchers with an invaluable opportunity to monitor the passengers' station-level demand (Nassir et al., 2015) and incorporate the derived insights into the formulation (Ingvardson et al., 2018). Yet, the application of this data requires additional analysis, especially employing advanced data-driven techniques to find meaningful patterns.

Aiming at improving passengers' travel time, our present interest lies in enhancing the performance of urban rail transit lines by employing the stop-skipping strategy at the operational level to address the unbalanced demand. Using a Long Short-Term Memory model, we seek to integrate the insights from the smart card data into an optimization stop-skipping model to reduce patrons' total travel time. The remainder of this paper is organized as follows. In Section 2, we provide a general overview of the literature of LSTM models and stop/skip strategy as well as our contribution. In Section 3, we discuss and visualize the data. In section 4, the proposed methodology including a deep learning model to predict the peak hour demand, an optimization model to find the stop/skip pattern, and a heuristic algorithm to solve the model are presented. Section 5 encompasses the results. Finally, in Section 6 we summarize and conclude the paper and provide future research directions.

## 2. Lessons from The Literature

### 2.1. Long Short-Term Memory Model

Long Short-Term Memory (LSTM) is a type of deep neural network (RNN) model with a powerful performance on time-series data due to its sophisticated structure to memorize the long-term historical data (Bianchi et al., 2017). Accordingly, in LSTM models, vanishing and exploding gradients can be reduced significantly in comparison to the other models such as general recurrent neural networks. Due to their great performance, LSTM models have recently been employed widely in different fields of research such as image captioning (Sharma and Jalal, 2020) and weather forecasting (Karevan and Suykens, 2020). In transportation engineering, researchers also have achieved great results by applying LSTM models in different research areas to predict and detect traffic accidents (Parsa et al., 2019), predict the delay in train operation (Huang et al., 2020), and arrival time of public transit modes (Liu et al., 2020). Investigating the performance of LSTM models with time-series data, Siami-Namini and Namin, (2018) compared LSTM with traditional techniques such as ARIMA and achieved more than 84% reduction in the error rates.

In studies focusing on the operation and performance of the urban transit systems, the recent application of deep learning techniques such as LSTM has been taken into consideration (Li et al., 2017). In this regard, researchers have attempted to take advantage of the capabilities of the LSTM model in coping with time-series data and predict the short-term subway ridership (Chen et al., 2020), predict passenger flow (Zhang et al., 2020), and schedule subway timetable (Kuppusamy et al., 2020). In 2020, Liu et al., compared LSTM with widely used methods such as random forest, deep neural network, recurrent neural network, and gated recurrent unit to predict hourly subway passenger flow and showed outperformance of LSTM against the other techniques. In another study, using an LSTM model, passenger flow of subway stations is predicted every 10 minutes and the promising results of the LSTM model over other techniques such as random forest confirms its capability to capture the linear and nonlinear relations in time-series data (Lin and Tian, 2020).

Despite the considerable performance of the LSTM model, integration of this powerful technique with urban transit line optimization models in a real-time framework is neglected in most of the related studies. To this end, this study is set out to bridge between the insights derived from performing a powerful deep learning technique on smart card data and the flexibility obtained by adopting the operational control stop-skipping strategy to cope with unbalanced demand rates in different stations.

### 2.2. Stop-Skipping Strategy and Contribution

Stop skipping (or limited-stop) strategy is a control technique that allows transit vehicles to skip certain stations along the path which can be devised at either tactical planning level or operational level. The objective at the tactical planning level is to address the disruptions occurring during actual operations while maintaining the good performance of the service (Gkiotsalitis and Cats, 2021). At this level, the pattern of stops and skips are determined at the start of the daily operation which is inflexible and cannot change during the day (Furth, 1986). On the other hand, at the operational level, the stop/skip scheme is devised before dispatching the vehicles from the terminus which accentuates the importance of finding computational methods in near real-time to solve the problem (Gkiotsalitis and Cats, 2021).

The flexibility and potential embedded in the stop skipping strategy to improve the efficiency of transit systems has been the source of inspiration for many studies. In this regard, scholars have introduced different stop strategies including zonal operation, express/local operation, and combined stop operation to diminish the total travel time compared to the all-stop scenario (Qi et al., 2018). The common theme in most research conducted in this area includes employing optimization models to improve the quality of service (Chew et al., 2019; Leiva et al., 2010; Qi et al., 2018). Developing a mathematical model with the objective of minimizing the costs of both passengers and operation agencies, Chen et al., (2015) analyzed the effects of applying stop-skipping on optimal headways, and substantiated the suitability of this strategy for headway between 3.5 and 5.5 minutes. Zhang et al., (2017) proposed a Flexible Skip-Stop Scheme (FSSS) for bi-directional metro operation during off-peak, which consists of a Mixed Integer Linear Programming (MILP), and genetic algorithm based solution. For the case of Shenzhen bi-directional metro, the results showed the reduction of average travel time by 50 seconds per passenger. Wang et al., (2014) considered the origin-destination (O-D) dependent train scheduling problem along with the stop-skipping strategy. At their approach, first, the optimization model with O-D rates is formulated. Then, a bi-level solution approach is proposed to solve the problem, in which the higher level employs a genetic algorithm for integer programming and the lower level applies a sequential quadratic programming approach. The simulation results show that the bi-level optimization approach can be applied to small-sized train lines.

Stop skipping also has been investigated as a means of providing equity-oriented services (Jomehpour Chahar Aman and Smith-Colin, 2020). In this respect, the urban rail transit network can be described as a distributed system, in which users are sharing a set of resources (i.e., train capacity) (Shang et al., 2018). Therefore, equitable individuals and groups demand a service operation that can offer equal shares of resources. This is especially critical in the presence of oversaturation when different users waiting at different stations are prone to varying shares of train capacity leading to the inequity problem under train conventional all-stopping pattern (Shang et al., 2018). Incorporating the stop skipping method, Shang et al., (2018) developed a linear programming formulation by introducing an equity index associated with the maximum number of missed trains of all passengers. Then, using a simple and a real-world case, they demonstrated the effectiveness of the stop and skip method to improve the system-wide equity performance.

Considering users whose destination is going to be skipped, providing sufficient information to remind passengers about the skipped stations of the coming train is of crucial importance. For implementation purposes, transit providers can employ electronic signs equipped at station platforms and travel information published in the APP on smartphones or Websites to inform users about the skip-stopping pattern (Shang et al., 2018).

With the goal of reducing patrons' travel times and increasing overall transit operation efficiency, the present research is set out to introduce a real-time data-driven framework to find the optimal stop and skip pattern for urban transit lines. We use the smart card data including both origin and destination stations of each passenger and their corresponding transaction time for an entire month to train a deep learning (LSTM) model. Employing the hourly demand rates from four preceding hours, this model can predict the station-level demand rates for the peak hour. Since true demand rates are only determined after all passengers have alighted at their destination station (i.e., after the peak hour), the significance of this model is that it provides the demand prior to the actual peak hour for each day. Given this predicted demand, we can move toward developing a method that accounts for the heterogeneity of day-to-day passengers' distribution. Then, the predicted rates, are inputted into a mathematical model with the objective of minimizing total passengers' travel times. Next, a heuristic algorithm is introduced to solve the minimization stop-skipping problem and reach the optimal pattern. Finally, the results are compared to the all-stop scenario to evaluate the models and the solution. The conceptual proposed framework can be seen in the flowchart illustrated in Figure 1.

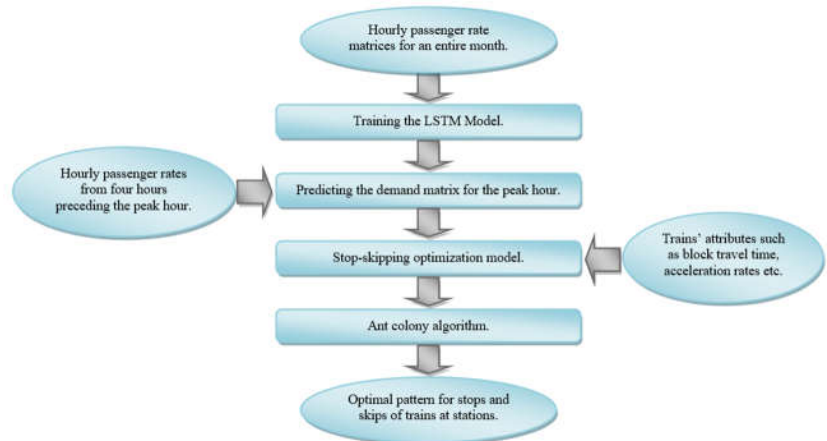

*Figure 1. The proposed framework to address the stop-skipping problem.*

## 3. Data

Figure 2 illustrates the Tehran urban rail network including 7 lines and 128 stations with the capacity of transporting over 1.2 million passengers per day. This overcrowded network is one of the busiest urban rail networks in the middle east making it a good candidate to test the stop-skipping strategy. In this study, we utilize the data of hourly trip rates between stations of line 1 (i.e., the red line in figure 2) which is extracted from the automated fare collection system. Comprised of 30 stations (including the terminus), this is the longest as well as the busiest line. In this study, we only considered the north to the south direction. The dataset includes 8,272,317 trips from 5:00 a.m. until 11:00 p.m. recorded from April 21, 2018 to May 21, 2018. The data includes a unique id number for each passenger, the time and the station of the transaction, and the type of the transaction (i.e., entry or exit).

The dispatching headway of trains from the terminus for the hour of the study is 5 minutes. Each train has the capacity of transporting 1,348 passengers. The running speed, acceleration and deceleration rates of trains are 19.44 m/s, 0.7 m/s$^2$, and -0.7 m/s$^2$, respectively.

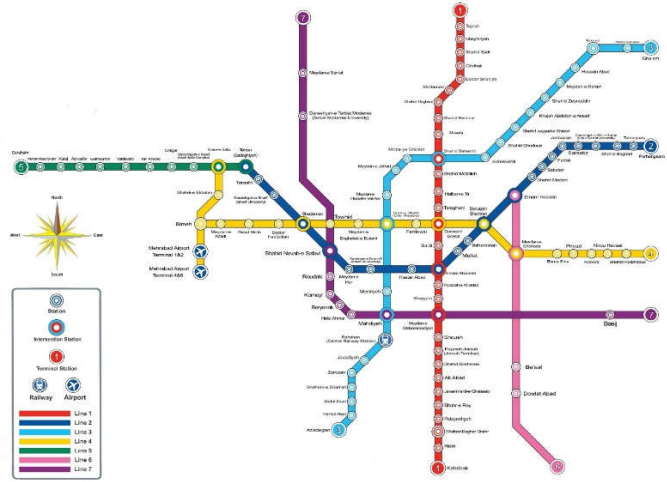

Figure 2. Tehran urban rail network [1].

To have a better understanding of demand between stations, hourly trips from 5:00 a.m. to midnight of May 13, 2018, that are generated from and attracted to all 30 stations of the red line are plotted in figure 3.

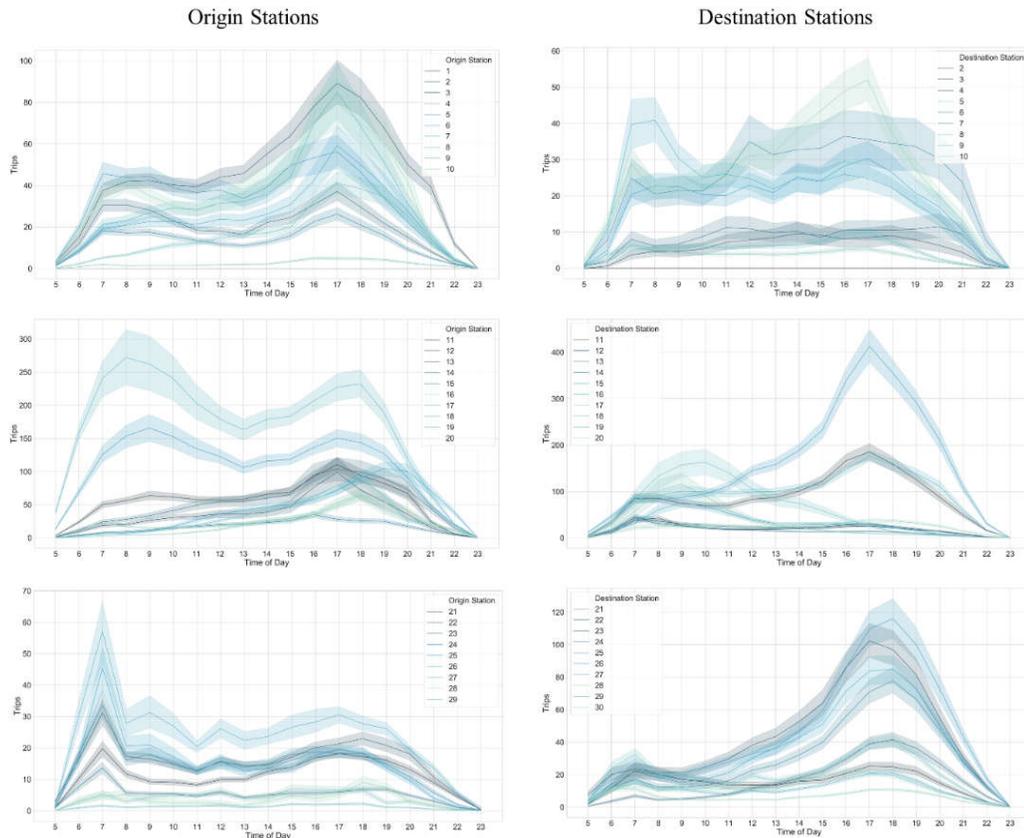

Figure 3. Hourly trips from 5:00 a.m. to midnight of May 13, 2018, that are generated/ attracted from/to 30 stations of line 1.

The three plots on the left side, represent the number of trips vs time of the day for the trips that start from stations 1-10, 11-20, and 21-29, respectively. Similarly, the three plots on the right side, represent the number of trips vs time of the day for the trips that end at stations 1-10, 11-20, and 21-29, respectively. According to this figure, the high demand for trips is shifting between morning and evening peak hours depending on the location of stations. Since this line connects the northern side of the city to the southern suburban areas, evening peak hours demand can be observed obviously in the trips generated from first stations in the north side and destinated to the final stations on the south side (i.e., home location of people who are returning home from work in the evening).

In this study, we selected the evening peak hour (i.e., 5:00 pm or 17:00) as the target time of day, for which, our proposed real-time stop-skipping approach can be implemented. To this end, to predict the station-level demand matrix for this time of day in real-time, we take advantage of the time-series demand data from 12:00 p.m. to 4:00 p.m. of the entire month in the procedure of training the deep learning model. The final data that we used for training the model includes 1,2180 observations. Besides the data from May 13th, 2018 is employed to test the performance of our real-time prediction model in our proposed stop-skipping approach.

## 4. Methodology

This section is comprised of three parts. Once the LSTM deep learning model is trained, we can predict station-to-station passenger rates for every desired hour based on four preceding hours. Here, we first introduce the employed LSTM model. Then, we move to elaborate on the optimization stop-skipping model. Finally, we discuss the solution algorithm to solve the minimization problem.

### *4.1. Long Short-Term Memory Model*

LSTM is a type of recurrent neural network model with the ability to overcome the vanishing or exploding gradient, which most standard RNN models are suffering from, through its powerful demand ability to process gradients. LSTM is comprised of cells which are internal processing units in LSTM, and gates that are keeping or overriding information in the memory cell, forgetting previous information and deciding how to access the memory cell (Hochreiter and Schmidhuber, 1997). The LSTM consists of three gates: forget gate, input gate, and output gate, each of which has a specific role in an LSTM unit. Forget gate determines the part of the information that needs to be forgotten and eliminated from the previous cell state (i.e., $h[t-1]$), the input gate decides how much of the new state (i.e., $h[t]$) should be updated, and the output gate determines the portion of the state that must be outputted. The main flow of information happens through the cell state. The cell state is updated in the forward process and the output is computed as displayed in the equations 1-6 (Bianchi et al., 2017):

forget gate: $\quad \sigma_f[t] = \sigma\,(W_f \cdot x[t] + R_f \cdot y[t-1] + b_f)$ (1)

candidate state: $\quad \tilde{h}[t] = g_1(W_h \cdot x[t] + R_h \cdot y[t-1] + b_h)$ (2)

input gate: $\quad \sigma_u[t] = \sigma\,(W_u \cdot x[t] + R_u \cdot y[t-1] + b_u)$ (3)

cell state: $\quad h[t] = \sigma_u[t]\,(\cdot)\,\tilde{h}[t] + \sigma_f[t]\,(\cdot)\,h[t-1]$ (4)

output gate: $\quad \sigma_o[t] = \sigma\,(W_o \cdot x[t] + R_o \cdot y[t-1] + b_o)$ (5)

output: $\quad y[t] = \sigma_o[t]\,(\cdot)\,g_2(h[t])$ (6)

Where $x[t]$ is the input at time $t$, $\sigma_o(\cdot)$ is a sigmoid function, $g_1(.)$ and $g_2(.)$ denote the pointwise nonlinear activation function, $(\cdot)$ denotes the entry wise multiplication between two vectors, $R_o$, $R_u$, $R_h$, and $R_f$ represents weight matrices of the recurrent connections, $W_o$, $W_u$, $W_h$, and $W_f$ are weight matrices for the inputs of LSTM cells, $b_o$, $b_u$, $b_f$, and $b_h$ are bias vectors (Bianchi et al., 2017). The LSTM model was developed using Python (v3.7.3) through the Koras (v2.2.4) Deep Learning Library that uses TensorFlow (v2.0.0b0) in the backend.

### *4.2. Stop-Skipping Optimization Model*

The introduced optimization model in this section is a modified version of the model proposed by Wang et al., (2014), making it not only more practical to implement but also easier to solve. In the model introduced in (Y. Wang et al., 2014), which is referred to as the *base model* in the following sections, decision variables include trains arrival time into stations, trains departure time from stations, trains running speed between stations, and a set of binary variables denoted by $y_{ij}$, representing the decision of train $i$ to stop/skip at the station $j$ for all trains and stations. Accordingly, the departure headway of the trains from the terminus also becomes a set of variable values.

Although the base model might achieve lower values for its objective function, it is likely to induce inconvenience for transit agencies and passengers. For example, in our study line, including 30 stations and manually driven trains, determining 29 different values for running speeds on 29 blocks (i.e., the distance between two consecutive stations) and expecting drivers to adhere to them, in the less than 2-min block travel times, does not seem practical. This is especially important knowing that for any control strategy to be successful, it is required to be acted upon by drivers (Gkiotsalitis and Cats, 2021). Besides, from the passengers' perspective, reliability is one of the most important advantages of transit systems, especially urban rail transit lines, implying that a fixed dispatching headway is likely to assist passengers to have better planning by reducing concerns about unpredictable schedule changes.

Moreover, to solve the base model, Wang et al., (2014) proposed a bilevel optimization approach, in which the upper level employs a heuristic algorithm to optimize the set of binary variables. Then, fixing the binary variables, the lower level uses sequential quadratic programming to optimize the other decision variables (i.e., running times and arrival/departure times to/from stations). In this regard, the computational time for an example comprised of 5 trains and 6 stations is reported approximately 37 minutes on a 1.8 GHz Intel Core 2 Duo CPU. Considering the size of real problems, like our study line including 30 stations along the path, solving the base model engenders a prohibitive amount of computational time.

Finally, the biggest concern with the base model is that it assumes fix station-to-station passenger rates which gives rise to two major issues:

I. Considering that the passenger rates pattern is subject to significant variations, depending on the time of the day as well as the day of the week, assuming a fixed matrix for passenger rates as the most important input to the optimization model can simply lead to fallacious results. To address this problem, we use the smart card data to extract the station-to-station passenger rates for each hour to account for the heterogeneity in passengers' distribution at different times.

II. Station-to-station passenger rate is posterior information, meaning that we first need to wait until passengers' trips are finished to determine both their origin and destination stations for each hour, and then input that to the optimization model (in the base model). Whereas, in practice, we need to establish the best stop/skip pattern before observing the finalized trips. To address this issue, in this study, we employ the deep learning LSTM model to predict the passenger rates of the peak hour based on the time-series data from four hours preceding the peak hour.

In the improved model presented in this study, we also strive to relax some assumptions of the base model, making it more tractable both in the implementation and the solution approach for real-size problems. Accordingly, regarding the movement of trains, we assume fixed running speeds in all blocks along the urban rail transit line. In other words, regarding the three phases in the movement of the train between stations (i.e., the acceleration phase, the deceleration phase, and the speed holding phase), we consider an average travel time for each block accounting for all of the three phases of movement. In this sense, the average travel time on block $j$ (i.e., $r_j$) can be calculated using the acceleration/deceleration rates and the constant speed of the train in the speed holding phase. Moreover, we assume a predefined constant dispatching headway from the terminus for all trains which is the conventional approach in most transit networks.

Incorporating the abovementioned assumptions into the formulation presented in Wang et al., (2014), we introduce a modified optimization model to determine the optimal pattern of stops and skips in urban transit lines to minimize total users' travel time. Since in this model, the only unknowns are the binary variables (i.e., $y_{ij}$), representing the decision of trains to stop at or to skip stations, our model can be solved at one level resulting in a significant computational simplicity. Overall, the following assumptions are embedded in the formulation regarding the trains and stations:

1. One direction of a line including $J$ stations is considered.
2. Trains cannot skip transfer stations with other lines.
3. Trains cannot pass each other at stations.
4. The values of travel times between every two stations, stop-to-stop passenger travel rates, acceleration/deceleration rates, and departure headway from the first station are inputs to the optimization model.
5. Each passenger uses only one train to get to his/her destination, and transfer between trains is not allowed.

Assumption 4 states that the stop-to-stop passenger rates are inputs to the optimization model. To clarify, it should be noted that although passenger rates are a given constant input to the optimization model, it is the output of the LSTM model (i.e., the prediction for the peak hour). Therefore, for each day, we can run the optimization model by utilizing the corresponding predicted passenger rates extracted from the LSTM model to account for the variations in passenger rates across different times.

The dynamic of the train movement at each station includes entering the station, dwelling at the station, and departing from the station. In the case of skipping a station, instead of deceleration, stop, and acceleration, the train continues its movement with the constant running speed. Accordingly, the train can save the amount of time equal to the difference between stopping at or skipping a station. In this regard, as an example, consider a train in the Tehran urban rail network with the constant holding speed, acceleration, and deceleration rates of 70 km/h (i.e., 19.44 m/s), 0.7 m/s$^2$, and -0.7 m/s$^2$, respectively. Using the simple kinematic physics formulas, the time required for the train to decrease its velocity from 70 km/h to zero (i.e., deceleration phase) equals 27.7 sec. On the other hand, in case of skipping the station, the train can travel the same distance with a constant speed of 70 km/h which only needs 13.88 sec. Accordingly, the train saves 13.82 sec (i.e., 27.7-13.88) by skipping the deceleration phase. Similarly, these values are identical for the acceleration phase, which results in a saving of 13.82 sec in the acceleration phase as well. Hereafter, we denote the summation of these two values (i.e., 13.82 +13.82) by $t^{acc}$. Moreover, the minimum dwell time of a train at a station is 25 sec based on the criteria established by the urban rail operator (i.e., $s_{criteria}$). Altogether, by skipping a station, the train saves 52.64 sec (i.e., 13.82+25+13.82) due to eliminating the deceleration, stop, and acceleration phases.

### 4.2.1. Model Formulation

Noticing that we refer to the link between stations $j$ and $j + 1$ as the block $j$, similar to (Y. Wang et al., 2014 and Yihui Wang et al., 2014) the following notations are used in the model :

| Symbol | Description | Symbol | Description |
|---|---|---|---|
| $j$ | Index of stations, from $1$ to $J$; | $i$ | Index of trains, from $1$ to $I$; |
| $r_j$ | Travel time on block $j$; | $a_{i,j}$ | Arrival time of train $i$ to station $j$; |
| $d_{i,j}$ | The departure time of train $i$ from station $j$; | $s_{i,j}$ | Dwell time of train $i$ at station $j$; |
| $t_{in-vehicle,i,j}$ | Passengers in-vehicle time for train $i$ on block $j$; | $t_{wait,i,j}$ | Passengers' waiting time for train $i$ at station $j$; |
| $n_{i,j}^{alight}$ | Number of passengers alighting from train $i$ at station $j$; | $n_{i,j}^{remain}$ | The remaining capacity of train $i$ at station $j$ immediately after the alighting process; |
| $n_{i,j}^{board}$ | Number of passengers boarding train $i$ at station $j$; | $w_{i,j}^{left}$ | Number of passengers left from train $i$ at station $j$ due to capacity limitation; |
| $n_{i,j}$ | Number of passengers on train $i$ immediately after departing from | $w_{i,j,k}^{wait}$ | Number of passengers waiting for train $i$ at station $j$ with destination $k$; |
| $w_{i,j}^{wait}$ | Number of total passengers waiting for train $i$ at station $j$; | $w_{i,j,k}$ | Number of passengers left from train $i$ at station $j$ with destination $k$; |
| $w_{i,j}$ | Number of total passengers left from train $i$ at station $j$; | $w_{i,j,k}^{want2}$ | Number of passengers at station $j$ with destination $k$ who want to board train $i$; |
| $u_{j,k}$ | The arrival rate of passengers with destination $k$ to station $j$; | $u_j$ | The total arrival rate of passengers to station $j$; |
| $h$ | Departure headway of trains from the first station (i.e., terminus); | $h_0$ | Minimum time between the arrival of two consecutive trains into a station; |
| $\alpha_1, \alpha_2, \alpha_3, \alpha_4, \gamma$ | Parameters; | $n^{door}$ | The number of doors of the train; |
| $C_{max}$ | The capacity of trains; | $y_{ij}$ | The binary variable signifying the stop (1) or skip (0) of train $i$ at station $j$; |
| $t_{in-vehicle,nom}$ | Total passengers' in-vehicle time in the all-stop scenario; | $t_{wait,nom}$ | Total passengers' waiting times in the all-stop scenario; |
| $s_{max}$ | Maximum dwelling time for the train (constant); | $t^{acc}$ | The time saved by eliminating acceleration and deceleration caused by skipping a |
| $s_{criteria}$ | The criteria dwell time established by the transit agency; | $N$ | The set of trains; |
| $w_I^{left}$ | Number of passengers left from the last train; | $M$ | The set of stations; |
| $w_{I,nom}^{left}$ | Number of passengers left from the last train in the all-stop scenario; and | $M^{transfer}$ | The set of transfer stations with other lines. |

Here, using the definitions presented in the base model, we try to briefly discuss the constraints. For a more detailed elaboration on the formulation please see (Y. Wang et al., 2014) and (Yihui Wang et al., 2014). The objective function (Eq. 7) is comprised of three normalized parts. The first two parts consider passengers' total in-vehicle and passengers' waiting times, respectively. The third part is a penalty term to deter the last train from leaving passengers behind.

Eq. 8 formulates the users' in-vehicle time for train $i$ on block $j$ including the running time on block $j$ and the waiting time of those who do not alight at station $j+1$. Eq. 9 describes passengers' waiting time at station $j$ for train $i$ including the waiting time of passengers left by the previous train as well as the newly arrived ones (Y. Wang et al., 2014).

$$\text{Min } Z = \frac{\sum_{i=1}^{I}\sum_{j=1}^{J-1} t_{in-vehicle,i,j}}{t_{in-vehicle,nom}} + \gamma \frac{\sum_{i=1}^{I}\sum_{j=1}^{J-1} t_{wait,i,j}}{t_{wait,nom}} + \frac{w_I^{left}}{w_{I,nom}^{left}} \tag{7}$$

$$t_{in-vehicle,i,j} = n_{i,j} \cdot t_j + \left(n_{i,j} - n_{i,j+1}^{alight}\right) \cdot s_{i,j+1} \cdot y_{i,j+1} \qquad \forall\, i,j \in (N,M) \tag{8}$$

$$t_{wait,i,j} = w_{i-1,j} \cdot (d_{i,j} - d_{i-1,j}) + \frac{1}{2} \cdot u_j \cdot (d_{i,j} - d_{i-1,j})^2 \qquad \forall\, i,j \in (N,M) \tag{9}$$

The constraints from Eq. 10 to Eq. 14 are related to the dynamic of the train movement. Eq. 10 is pertinent to vehicles' dispatching headway from the terminus which is a constant value. The arrival time of train $i$ to station $j+1$ equals the sum of the departure time of train $i$ from station $j$ and running time on block $j$ (Eq. 11). Eq. 12 gives the departure time of the train $i$ from station $j$ which depends upon the skipping decision on this station.

$$d_{i,1} = d_{i-1,1} + h \qquad \forall\, i \in N \tag{10}$$

$$a_{i,j+1} = d_{i,j} + t_j \qquad \forall\, i,j \in (N,M) \tag{11}$$

$$d_{i,j} = a_{i,j} + s_{i,j} \cdot y_{i,j} \qquad \forall\, i,j \in (N,M) \tag{12}$$

Eq. 13 addresses the dwell time at stations. In this regard, if train $i$ stops at station $j$ (i.e., $y_{i,j} = 1$), then dwell time depends on not only the number of boarding and alighting passengers but also on the number of waiting passengers at the platform (Bi-chun et al., 2011). Thus, first, we find the time required for boarding and alighting passengers and compare it to the dwell time announced by the transit agency. Then, we add the higher value to $t^{acc}$ (i.e., the saved time due to eliminating acceleration/deceleration) to account for the time saved by skipping a station. The ratio $w_{i,j}^{wait}/n^{door}$ in Eq. 13 indicates the number of waiting passengers in front of each vehicle's door assuming a uniform distribution of passengers across train doors. Addressing the safety of trains movement, constraint 14 defines a minimum interval ($h_0$) between the departure of a train from a station and the arrival of the subsequent train (Y. Wang et al., 2014).

$$s_{i,j} = \max\left(\alpha_1 + \alpha_2 n_{i,j}^{alight} + \alpha_3 n_{i,j}^{board} + \alpha_4 \left(\frac{w_{i,j}^{wait}}{n^{door}}\right)^3 n_{i,j}^{board}, s_{criteria}\right) + t^{acc} \tag{13}$$
$$\forall\, i,j \in (N,M)$$

$$a_{i,j} - d_{i-1,j} \geq h_0 \qquad \forall\, i,j \in (N,M) \tag{14}$$

The constraints from Eq. 15 to Eq. 27 are related to passenger train interaction. The number of passengers waiting for train $i$ at station $j$ with destination $k$ can be calculated by adding passengers left from the preceding train (i.e., $w_{i-1,j,k}$) to the newly arrived passengers between the departure of train $i-1$ and arrival of train $i$ (Eq. 16). Eq. 17 is the summation of waiting passengers over all destinations. Eq. 18 defines passengers who want to board trains at stations. Considering that transfer between trains is not permitted, the willingness of passengers with destination $k$ waiting at station $j$ to board train $i$ is contingent upon the stopping decision of the train $i$ in both stations $j$ and $k$. In other words, if train $i$ is planned to skip either origin or destination stations of a passenger, he/she does not want to board train $i$ (Eq. 19). The number of passengers on the train $i$ who want to alight at station $j$ equal the summation of passengers who have been boarded in previous stations having station $j$ as their destination (Eq. 20). Eq. 21 calculates the available capacity of train $i$ immediately after alighting passengers at station $j$ and before boarding new passengers. The number of passengers boarding train $i$ at station $j$ equals the minimum value between the available capacity after the alighting process and the number of users at station $j$ who want to board train $i$ (Eq. 22) (Y. Wang et al., 2014).

$$u_j = \sum_{k=j+1}^{J} u_{j,k} \qquad \forall j \in M \qquad (15)$$

$$w_{i,j,k}^{wait} = w_{i-1,j,k} + u_{j,k}(d_{i,j} - d_{i-1,j}) \qquad \forall i, j \in (N, M), k \in j+1, \ldots, J \qquad (16)$$

$$w_{i,j}^{wait} = \sum_{m=j+1}^{J} w_{i,j,m}^{wait} \qquad \forall i, j \in (N, M) \qquad (17)$$

$$w_{i,j,k}^{want2} = y_{i,j} \, y_{i,m} \, w_{i,j,k}^{wait} \qquad \forall i, j \in (N, M), k \in j+1, \ldots, J \qquad (18)$$

$$w_{i,j}^{want2} = y_{i,j}\left( w_{i,j,J}^{wait} + \sum_{k=j+1}^{J-1} y_{i,k} w_{i,j,k}^{wait} \right) \qquad \forall i, j \in (N, M) \qquad (19)$$

$$n_{i,j}^{alight} = \sum_{l=1}^{j-1} n_{i,l,j}^{board} \qquad \forall i, j \in (N, M) \qquad (20)$$

$$n_{i,j}^{remain} = C_{max} - n_{i,j-1} + n_{i,j}^{alight} \qquad \forall i, j \in (N, M) \qquad (21)$$

$$n_{i,j}^{board} = \min(n_{i,j}^{remain} - w_{i,j}^{want2}) \qquad \forall i, j \in (N, M) \qquad (22)$$

Eq. 23 and Eq. 24 determine the number of users left by train $i$ at station $j$. Depending on whether train $i$ stops at station $j$ or not, we have two situations (Y. Wang et al., 2014):

1. Train $i$ skips station $j$ (i.e., $y_{i,j} = 0$): no one at station $j$ can board train $i$.
2. Train $i$ stops at station $j$ (i.e., $y_{i,j} = 1$): if $w_{i,j}^{want2} < n_{i,j}^{remain}$, all passengers who want to board can get on train $i$. On the other hand, if $w_{i,j}^{want2} > n_{i,j}^{remain}$, some passengers cannot board train $i$. Passengers left from train $i$ at station $j$ due to capacity limitation ($w_{i,j}^{left}$) can be calculated using Eq. 23. Note that in this case, the assumption holds that the number of passengers with destination $k$ who are left by train $i$ is proportionate to the passengers at station $j$ who want to board:

$$w_{i,j,k} = w_{i,j}^{left} \frac{w_{i,j,k}^{want2}}{w_{i,j}^{want}} \qquad \text{if } y_{i,j} = 1 \text{ and } y_{i,k} = 1.$$

Taking these remarks into consideration, Eq. 24 gives the number of passengers left (either for the capacity limitation or skipping decisions) by train $i$ at station $j$ with destination station $k$ (Y. Wang et al., 2014). Eq. 25 represents the summation of passengers with all possible destinations left from train $i$ at station $j$.

$$w_{i,j}^{left} = y_{i,j}\left[w_{i,j}^{want2} - \min(n_{i,j}^{remain}, w_{i,j}^{want2})\right] \qquad \forall i, j \in (N, M) \qquad (23)$$

$$w_{i,j,k} = y_{i,j}\left(y_{j,k} w_{i,j}^{left} \frac{w_{i,j,k}^{want2}}{w_{i,j}^{want}} + (1 - y_{j,k})w_{i,j,k}^{wait}\right) + (1 - y_{i,j})w_{i,j,k}^{wait} \qquad (24)$$

$$\forall i, j \in (N, M), k \in j+1, \ldots, J$$

$$w_{i,j} = \sum_{k=j+1}^{J} w_{i,j,k} \qquad \forall i, j \in (N, M) \qquad (25)$$

By deducting the number of passengers left from train $i$ at station $j$ with destination $k$ from the waiting ones, the number of boarding passengers at each station can be calculated (Eq 26). The number of passengers inside the train in each block can be calculated by knowing the number of passengers in the previous block, and the number of alighting/boarding passengers in the previous station (Eq. 27) (Y. Wang et al., 2014).

$$n_{i,j,k}^{board} = w_{i,j,k}^{wait} - w_{i,j,k} \qquad \forall\, i,j \in (N,M),\, k \in j+1,\dots,J \qquad (26)$$

$$n_{i,j} = n_{i,j-1} - n_{i,j}^{alight} + n_{i,j}^{board} \qquad \forall\, i,j \in (N,M) \qquad (27)$$

Eq. 28 ensures that no two successive trains skip the same station. Finally, the last constraint expresses no skipping is allowed at transfer stations with other lines.

$$y_{i,j} + y_{i+1,j} \geq 1 \qquad \forall\, i,j \in (N,M) \qquad (28)$$

$$y_{i,j} = 1 \qquad \forall\, i \in N,\, j \in M^{transfer} \qquad (29)$$

Concerning the presented model, the following notes should be mentioned:
1. Eq. 10 is added to the base model to account for the fix dispatching headway.
2. The average travel time in each block is incorporated into the formulation via Eq. 11.
3. Eq. 28 is added to prevent the inconvenience caused by two successive trains skipping the same station.
4. Eq. 29 is added to prevent trains from skipping transfer stations with other lines.
5. After determining the only decision variables of the model (i.e., the set of $y_{i,j}$), all other variables such as arrival/departure times to/from stations can be found (as opposed to the base model in which arrival/departure times are also decision variables). This makes it feasible to solve the model in only one level (rather than the bilevel nature of the base model), acting as a significant contributor to the computational simplicity.

### 4.3. Solution Algorithm

In the optimization model introduced in section 4.2.1, the number of binary decision variables is the paramount factor determining the computational time. Therefore, in real problems, the exponential nature of the problem translates into a prohibitive amount of calculation. Here, we introduce a heuristic algorithm based on ant colony optimization (ACO) (Dorigo et al., 2006) to employ swarm intelligence and search for better solutions.

In the stop-skipping model, the number of decision variables equal to the multiplication of trains by stations. To utilize the concept of ACO, we can consider each decision variable as a layer as can be seen in figure 4. Each layer also has two nodes including 0 and 1, representing the decision to stop or skip for the corresponding binary variable. We consider a virtual ant entering to the first layer corresponding to the decision variable (i.e., $y_{11}$). In this layer, the ant has two node options to choose from. The probability of choosing either of the nodes is calculated from Eq. 30. Here, $j$ is the index of the node $\in \{0,1\}$, $i$ is the index of layer, $k$ is the index of ant, $P_{ii}^k$ is the probability of selecting node $j$ from layer $i$ by ant $k$, $\zeta_{ij}$ is the pheromone on node $j$ from layer $i$, and $\alpha$ is a parameter.

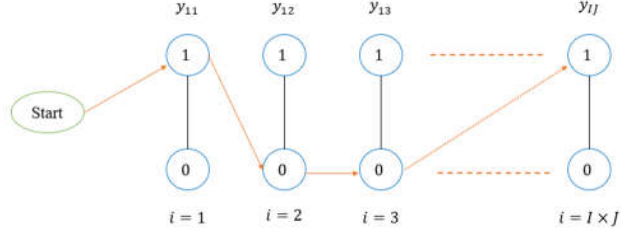

*Figure 4. Movement of ants across layers.*

$$P_{ij}^k = \frac{(\zeta_{ij})^\alpha}{\sum_j (\zeta_{ij})^\alpha} \quad \forall j \in 0,1 \ \& \ i = 1,2,..I \times J \ \& \ k = 1,2,.. Number\ of\ Ants \tag{30}$$

$$\sum_j P_{ij}^k = 1 \quad \forall k \in nAnt \ \& \ \forall i \in \{1,2,..I \times J\} \tag{31}$$

In Eq. 30, $\zeta_{ij}$ is the amount of pheromone on node $j$ in layer $i$ representing the information collected from the social intelligence of all ants. Besides, Eq. 31 guarantees that the summation of the probabilities of choosing a node within a layer equals one.

### 4.3.1. Algorithm Steps

**Step 1**. Initialization: set the number of ants ($nAnt$), number of variables or layers ($nVar$), number of iterations ($maxIt$), α coefficient, initial pheromone ($Q$), evaporation rate ($\rho$), and an appropriate value for pheromones on nodes ($\zeta_{ij}^0$) proportionate to the initial pheromone ($Q$).

- **Step 1-1.** Create a structure for each ant (i.e., $k \in nAnt$) including two parts: The basket of the ant (i.e., $TourAnt(k)$) containing the corresponding solution of each ant and the cost of the ant (i.e., CostAnt(k)) containing the value of the objective function in the stop-skipping model for its corresponding basket. Then, set the $CostAnt(k) = TourAnt(k) = \emptyset$.
- **Step 1-2.** Create two structures: one for the best ant of each iteration (i.e., $BestAnt(It)$) and another one for the best ant of all previous iterations (i.e., $BestAnt$):set: $CostBestAnt = CostBestAnt(It) = inf$, and $TourBestAnt = TourBestAnt(It) = \emptyset$.

**Step 2.** The main loop: set $It = 1$.

- **Step 2-1.** Set the index of the first ant: $k=1$.
- **Step 2-2.** Set the index of the first layer: $i = 1$.
  - **Step 2-2-1.** Using equation 30, calculate the probability of choosing each of two nodes in the current layer $i$.
  - **Step 2-2-2.** Utilizing the Roulette wheel method, choose one of the nodes within the layer $i$, then add that to the $TourAnt\ (k)$.
  - **Step 2-2-3.** If $i < nVar$, then set $i = i + 1$ and go to the step 2-2-1. Otherwise, go to the step 2-3.
- **Step 2-3.** Compute the value of the objective function of the train stop-skipping model with the binary variables within the $TourAnt\ (k)$. Then, put this value in $CostAnt(k)$.
  - **Step 2-3-1.** If $CostAnt(k) < CostBestAnt(It),$ consider the current $ant\ (k)$ as the best ant of the current iteration: $BestAnt(It) = Ant(k)$.
- **Step 2-4.** If $k < nAnt$, set $k = k + 1$ and go to the step 2-2. Otherwise, go to the step 2-5.

- **Step 2-5.** If $CostBestAnt(It) < CostBestAnt$, consider the best ant of the iteration as the best ant of all iterations: $BestAnt = BestAnt(It)$.

**Step 3.** Update: Using Eq. 32 update the pheromones of the nodes inside the baskets of the best ant of the current iteration ($BestAnt(It)$) as well as the best ant of all the previous iterations ($BestAnt$).

$$\Delta_{ij}^k = \begin{cases} \frac{Q}{Z(\psi^k)} & \forall j \in (\psi^+, \psi^*) \ \& \ \forall k \in (k^+, k^*) \ \& \ \forall i = 1 \ldots, I \times J \\ 0 & \text{Otherwise.} \end{cases} \quad (32)$$

Here, $\Delta_{ij}^k$ is the amount of pheromone that ant $k$ releases on node $j$ from layer $i$. $k^+$ and $k^*$ are the best ant of the current iteration and the best ant of all previous iterations, respectively. $\psi^+$ and $\psi^*$ are the set of traveled nodes by the best ant of the current iteration and the best ant of all previous iterations, respectively. $Q$ is the amount of the initial pheromone and $Z(\psi^k)$ is the value of the objective function of the stop-skipping model for the solution of the ant $k$.

**Step 4.** Evaporation: using equation (33) decrease the pheromone of all nodes.

$$\zeta_{ij} = (1 - \rho) \times \zeta_{ij} \quad (33)$$

**Step 5.** Stop criterion: If $It < maxIt$, set $It = It + 1$ and go to step 2-1. Otherwise, go to step 6.
**Step 6.** End: Consider the basket of the best ant among all iterations (i.e., $TourBestAnt$) as the final solution.

## 5. RESULTS
### 5.1. Peak Hour Station-Level Demand Prediction

We trained and tuned an LSTM model to predict the demand between all possible pairs of 30 stations in real-time for evening peak hour. To this end, we used the demand data between stations for four consecutive hours from 12:00 pm to 4:00 pm as the input of the LSTM model to predict the demand for the evening peak hour from 5:00 pm to 6:00 pm. To train the model, data is split randomly into 70% and 30% for training and validating the model, respectively.

Similar to other deep learning models, LSTM has many parameters (i.e., model weights), and hyperparameters (i.e., model parameters) that should be calculated and tuned in the training procedure. Various approaches can be applied to update the parameters and we used the Adam optimizer in this study. Regarding hyperparameter tuning, we applied the cross-validation technique to select the hyperparameters of the LSTM model. Therefore, the losses were computed through the Mean Squared Error regression loss class, the hidden layers of the model consisted of the Relu activation function, and in the output layer, the sigmoid activation function is utilized. Besides, the value of batch size, number of epochs, and learning rates are selected as 35, 500, and 0.001 respectively, through the hyperparameter tuning process. Figure 5 (a) displays the loss against epoch plot for training and validation data of the LSTM model demonstrating the good performance of this model.

After training the LSTM model, its performance in the prediction of the demand between stations is tested for a specific day that was not used in the training procedure (i.e., May 13th, 2018). Figure 5 (b) plots predicted value against true value for this model. The LSTM model achieves an accuracy of 97.08% for predicting the demand between stations for the period of 5:00 pm to 6:00 pm (i.e., the peak hour).

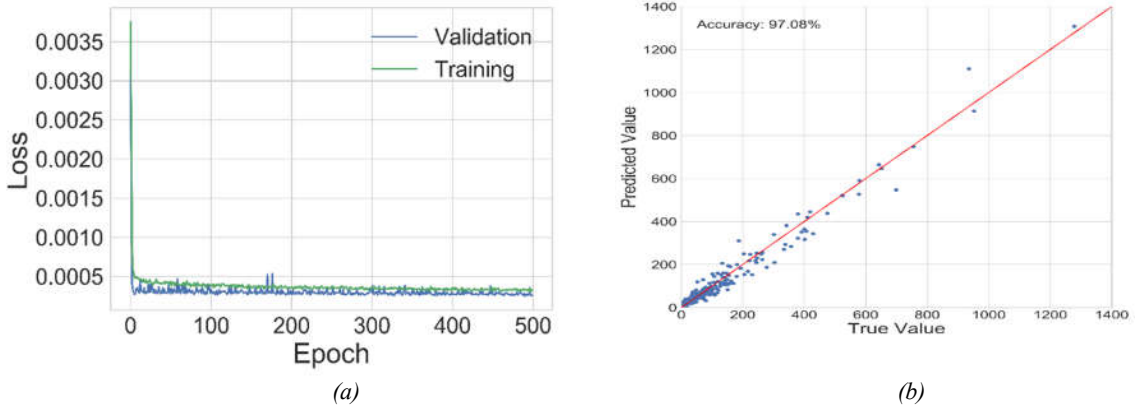

*(a)* *(b)*

*Figure 5. The loss against epoch plot for the LSTM model (a). Prediction accuracy of the LSTM model (b).*

### 5.2. Stop-skipping Model Results

Incorporating the predicted station-level demand rates provided by the LSTM model as an input to the optimization stop-skipping model, here, we analyze the performance of the proposed solution algorithm to investigate the results of applying the stop-skipping strategy on the southbound direction of our study line.

Regarding the weight of the second term of the objective function, $\gamma = 2$ was selected to account for the more negative perception of the waiting time, compare to the in-vehicle time, by users of the transit network (Sun and Hickman, 2005).

Besides, a combination of literature suggestions (see, for example, (Kucukkoc et al., 2013) and (Ahn and Ramakrishna, 2002)), as well as numerous times of try and error, were checked to find the best values for calibrating the solution approach which is shown in table 1.

*Table 1. Parameters of the proposed ACO algorithm.*

| Evaporation rate ($\rho$) | 0.1 | Number of layers ($nAnt$) | 360 |
|---|---|---|---|
| Parameter ($\alpha$) | 0.8 | Number of iterations ($maxIt$) | 30 |
| Initial pheromone ($Q$) | 7 | | |

Employing these parameters, *predicted* demand rates from section 5.1, and the operational data of the rail transit system presented in the Data section, we applied the proposed ACO algorithm to solve the stop-skipping model for 12 trains and 30 stations. The result is the stop and skips pattern during the peak hour presented in table 2. For the sake of compactness, only the stations that had at least one train skipping them, are presented in table 5. This table indicates that, for instance, train 4 skips stations 9 whereas train 5 stops at all stations. Here, we only have stations 2, 3, 4, 6, 8, 9, 10, 21, 23, 24, 27, and 29, implying that all other stations, which have not been mentioned in this table, are not allowed to be skipped by any trains. Trains 1, 5, and 12 should stop at all stations, and in total, 15 zeros exist in this table suggesting that, in the movement of 12 trains along 30 stations in the peak hour, 15 times skipping occurs. According to the result, approximately 47% of the total skipping decisions (i.e., 7 times) happen on the first 8 stations, 33% (i.e., 5 times) on the last 8 stations, and 20% (i.e., 3 times) on the 14 intermediate stations. Therefore, stations located at the two ends of the line (especially at the start section) are more prone to be skipped by trains under the optimality. This can be attributed to the lower number of passengers who originated/destined from/to these stations compared to the intermediate ones. This implies that the amount of time and available train capacity saved from skipping those stations provides the opportunity to lower total passengers' travel times and establish a demand-responsive transit service by assigning the trains to where there are demanded the most.

*Table 2. The result of solving the stop-skipping model using the proposed ACO for the study line.*

| Station  | 2 | 3 | 4 | 6 | 8 | 9 | 10 | 21 | 23 | 24 | 27 | 29 |
|----------|---|---|---|---|---|---|----|----|----|----|----|----|
| Train 1  | 1 | 1 | 1 | 1 | 1 | 1 | 1  | 1  | 1  | 1  | 1  | 1  |
| Train 2  | 1 | 1 | 1 | 1 | 1 | 1 | 1  | 1  | 0  | 1  | 1  | 1  |
| Train 3  | 1 | 1 | 1 | 1 | 0 | 1 | 1  | 1  | 1  | 1  | 1  | 1  |
| Train 4  | 1 | 1 | 1 | 1 | 1 | 0 | 1  | 1  | 1  | 1  | 1  | 1  |
| Train 5  | 1 | 1 | 1 | 1 | 1 | 1 | 1  | 1  | 1  | 1  | 1  | 1  |
| Train 6  | 0 | 1 | 1 | 1 | 1 | 1 | 1  | 1  | 1  | 1  | 1  | 0  |
| Train 7  | 1 | 1 | 0 | 1 | 0 | 1 | 1  | 1  | 1  | 1  | 1  | 1  |
| Train 8  | 1 | 1 | 1 | 1 | 1 | 1 | 0  | 1  | 1  | 0  | 0  | 1  |
| Train 9  | 1 | 0 | 1 | 1 | 1 | 1 | 1  | 1  | 1  | 1  | 1  | 1  |
| Train 10 | 0 | 1 | 1 | 1 | 1 | 1 | 1  | 1  | 1  | 1  | 1  | 1  |
| Train 11 | 1 | 1 | 1 | 0 | 1 | 1 | 1  | 0  | 1  | 1  | 0  | 1  |
| Train 12 | 1 | 1 | 1 | 1 | 1 | 1 | 1  | 1  | 1  | 1  | 1  | 1  |

Moreover, based on the result, non of the transfer stations with other lines (i.e., stations 11,15,17, and 20) are skipped, and no two consecutive trains skip the same station to avoid causing inconvenience for the waiting passengers.

Using the *true* and *predicted* station-level demand rates, the results of solving the model are presented in table 3. The first column is obtained by comparing the result of solving the stop-skipping model based on the *true* (i.e., observed) demand rates for both all-stop and stop-and-skip strategies. On the other hand, the second column uses the *predicted* demand rates obtained from the LSTM model for both all-stop and stop-and-skip cases. Please note that, in reality, the true values are only known after the trains alight all passengers which makes them posterior information. That is why we have strived to use the prediction of the LSTM model to have the demand rates before running the optimization model. Based on this table, the improvement in the objective function equals 3.55 and 4.01 percent for the true and predicted demand rates, respectively. This table also reveals that employing the predicted rates, the application of stop-skipping strategy instead of the all-stop approach has resulted in an improvement in both total passengers waiting time, by 4.65 percent as well as total passengers in-vehicle time, by 1.77 percent. These results, however, are slightly lower in the case of using the true rates. All, in all, these findings substantiate that this operational control has the potential to mitigate the overcrowded situations during the peak hour.

Considering the real-time application of the proposed framework in this study, the computational time to solve the model is of paramount importance. The convergence procedure of finding the result (i.e., table 2), is displayed in figure 6 (a) for the 30 iterations. It should be noted that the value of the objective function for the case of the all-stop scenario equals 4. At the first step, the ACO algorithm starts by generating random feasible solutions, sorting them, and finding the best ant which achieves the value of approximately 6.5 for the objective function. Then, at each iteration, the algorithm tries to employ the information from the previous ones and move toward a better solution. After 30 iterations, the value of the objective function reaches 3.839 equal to a 4.01% improvement.

Using a 2.70 GHz Intel (R) Core i5-72000U CPU running on a windows 10 operating system, the computational time is 107 seconds which is desirable considering the 300-sec headway between the departure of trains from terminus for the peak hour.

Table 3. *The effects of applying stop-skipping strategy on different parts of the objective function, using true and predicted station-level demand rates.*

|  | True demand rates | | Predicted demand rates | |
|---|---|---|---|---|
|  | All-stop strategy | Stop-skipping strategy | All-stop strategy | Stop-skipping strategy |
| Value of the objective function | 4 | 3.858 | 4 | 3.839 |
| Improvement in the objective function (%) | - | **3.55** | - | **4.01** |
| Value of the passengers' waiting time (per-hour) | 1680 | 1605 | 1854 | 1768 |
| Improvement in the passengers' waiting time (%) | - | **4.45** | - | **4.65** |
| Value of the passengers' in-vehicle time (per-hour) | 11164 | 11010 | 10917 | 10723 |
| Improvement in the passengers' in-vehicle time (%) | - | **1.38** | - | **1.77** |

Therefore, the proposed ACO-based solution algorithm can be employed to optimize the stop and skip pattern, which in fact, proves the tractability of the proposed framework to be implemented in a real-time context.

The schedule of 12 trains moving along 30 stations as well as the stop and skip pattern can be observed in figure 6 (b). Each line corresponds to an urban train and each specified pentagram in this figure represents a skipped station. Moreover, the slope of each line indicates the speed of the corresponding train at any time. As can be seen, this slope is equal on both sides of any skipped station, implying passing the station at a constant speed.

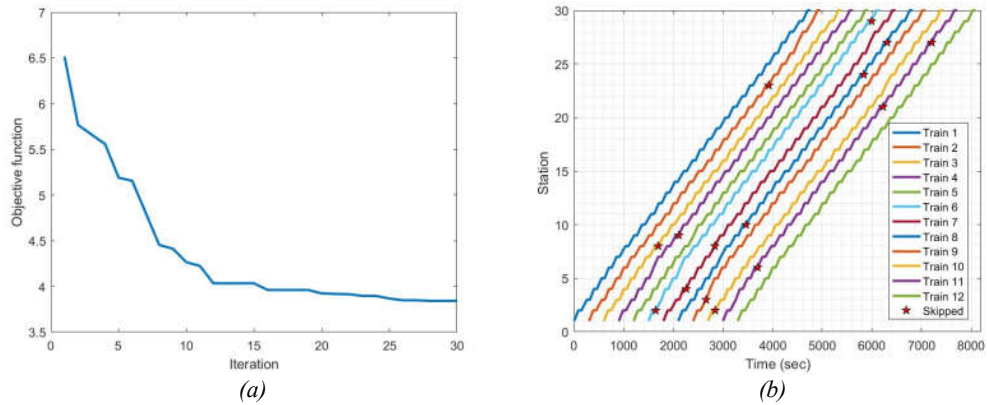

*(a)* *(b)*

Figure 6. *The convergence of the ACO solution algorithm in 30 iterations with the corresponding values of the objective function (a), and the schedule and the stop/skip pattern of trains for 12 trains and 30 stations (b).*

## 6. SUMMARY AND CONCLUSION

In this study, we proposed a data-driven approach to address the real-time stop-skipping problem through the integration of a deep learning prediction model into a stop-skipping optimization model. First, using the time-series smart card data of a real transit line, we trained an LSTM model with an accuracy of 97.08% to predict the station-level demand matrix for the evening peak hour. This model is capable of discovering linear and nonlinear relations in time-series data, and consequently providing a good prediction based on the four hours preceding the peak hour. This prediction allows us to have the passengers' demand rates, which is inherently posterior information that can be observed after all passengers reach their destinations, prior to the actual peak hour of every day. Then, we incorporated the output of the LSTM model as an input to an integer programming optimization model to minimize total passengers' travel times. The optimization model introduced in this study is an improved version of the model presented by Y. Wang et al., (2014) (i.e., the base model). The improved model is an integer programming problem with only binary decision variables corresponding to the decisions to stop or to skip at each station for all trains. This model can be solved at one level resulting in a significant computational simplicity which makes it feasible to apply on real-size lines including numerous trains and stations. Considering the exponential nature of the optimization model, a heuristic algorithm based on the Ant-Colony optimization was introduced. Using the information collected from the smart card data, we implemented the proposed stop-skipping approach in case of operating 12 trains traveling along 30 stations. The result indicates a 4.01% improvement in the objective function in comparison to the all-stop strategy. Accordingly, total passengers' in-vehicle time and total passengers' waiting time are improved by 1.77%, and 4.65%, respectively. The skipped stations in the optimized pattern are predominantly located in the two ends of the line, and more specifically at the first section of the line, suggesting that preserving the capacity of the trains for more crowded stations can lead to better exploitation of the existing resources which is beneficial for all passengers. To mitigate the effects on passengers whose stations are going to be skipped, we added constraints that no two consecutive trains are allowed to skip the same station, and also transfer stations can not be skipped. Plus, there is a penalty term in the objective function regarding the passengers that are left.

In general, the proposed method is more suitable for highly crowded transit lines with high stop-to-stop demand. In this study, we only considered one direction of a line. For future research purposes, all lines of an urban rail network along with the interdependency between different lines can be considered through a more complex framework. Moreover, analyzing the performance of the stop skipping strategy combined with other operational control measures such as vehicle holding and compare the results to the present study can be the next step for future works.

**Declaration of interests**

❑ The authors declare that they have no known competing financial interests or personal relationships that could have appeared to influence the work reported in this paper.

# 7. References

[1] [WWW Document], n.d. URL http://sevenhostels.com/images/MetroMap.jpg (accessed 1.11.21).

Ahn, C.W., Ramakrishna, R.S., 2002. A genetic algorithm for shortest path routing problem and the sizing of populations. IEEE Trans. Evol. Comput. 6, 566–579.

Asgharzadeh, M., Shafahi, Y., 2017. Real-Time Bus-Holding Control Strategy to Reduce Passenger Waiting Time. Transp. Res. Rec. J. Transp. Res. Board 2647, 9–16.

Bianchi, F.M., Maiorino, E., Kampffmeyer, M.C., Rizzi, A., Jenssen, R., 2017. Recurrent neural networks for short-term load forecasting: an overview and comparative analysis.

Chen, D., Zhang, J., Jiang, S., 2020. Forecasting the Short-Term Metro Ridership With Seasonal and Trend Decomposition Using Loess and LSTM Neural Networks. IEEE Access 8, 91181–91187.

Chen, J., Liu, Z., Zhu, S., Wang, W., 2015. Design of limited-stop bus service with capacity constraint and stochastic travel time. Transp. Res. Part E Logist. Transp. Rev. 83, 1–15.

Chen, X., Hellinga, B., Chang, C., Fu, L., 2015. Optimization of headways with stop-skipping control: a case study of bus rapid transit system. J. Adv. Transp. 49, 385–401.

Chew, J.S.C., Zhang, L., Gan, H.S., 2019. Optimizing limited-stop services with vehicle assignment. Transp. Res. Part E Logist. Transp. Rev. 129, 228–246.

Dai, Z., Liu, X.C., Chen, Z., Guo, R., Ma, X., 2019. A predictive headway-based bus-holding strategy with dynamic control point selection: A cooperative game theory approach. Transp. Res. Part B Methodol. 125, 29–51.

Ding, X., Guan, S., Sun, D.J., Jia, L., 2018. Short turning pattern for relieving metro congestion during peak hours: the substance coherence of Shanghai, China. Eur. Transp. Res. Rev. 10, 1–11.

Dorigo, M., Birattari, M., Stutzle, T., 2006. Ant colony optimization. IEEE Comput. Intell. Mag. 1, 28–39.

Fan, W., Ran, Y., 2021. Planning skip-stop services with schedule coordination. Transp. Res. Part E Logist. Transp. Rev. 145, 102119.

Furth, P.G., 1986. ZONAL ROUTE DESIGN FOR TRANSIT CORRIDORS. Transp. Sci. 20, 1–12.

Gkiotsalitis, K., 2019. Robust Stop-Skipping at the Tactical Planning Stage with Evolutionary Optimization. Transp. Res. Rec. J. Transp. Res. Board 2673, 611–623.

Gkiotsalitis, K., Cats, O., 2021. At-stop control measures in public transport: Literature review and research agenda. Transp. Res. Part E Logist. Transp. Rev. 145, 102176.

Hochreiter, S., Schmidhuber, J., 1997. Long short-term memory. Neural Comput. 9, 1735–1780.

Huang, P., Wen, C., Fu, L., Lessan, J., Jiang, C., Peng, Q., Xu, X., 2020. Modeling train operation as sequences: A study of delay prediction with operation and weather data. Transp. Res. Part E Logist. Transp. Rev. 141, 102022.

Ingvardson, J.B., Nielsen, O.A., Raveau, S., Nielsen, B.F., 2018. Passenger arrival and waiting time distributions dependent on train service frequency and station characteristics: A smart card data analysis. Transp. Res. Part C Emerg. Technol. 90, 292–306.

Jomehpour Chahar Aman, J., Smith-Colin, J., 2020. Transit Deserts: Equity analysis of public transit accessibility. J. Transp. Geogr. 89, 102869. https://doi.org/10.1016/J.JTRANGEO.2020.102869

Karevan, Z., Suykens, J.A.K., 2020. Transductive LSTM for time-series prediction: An application to weather forecasting. Neural Networks 125, 1–9.

Kucukkoc, I., Karaoglan, A.D., Yaman, R., 2013. Using response surface design to determine the optimal parameters of genetic algorithm and a case study. Int. J. Prod. Res. 51, 5039–5054.

Kuppusamy, P., Venkatraman, S., Rishikeshan, C.A., Reddy, Y.C.A.P., 2020. Deep learning based energy efficient optimal timetable rescheduling model for intelligent metro transportation systems. Phys. Commun. 101131.

Larrain, H., Muñoz, J.C., Giesen, R., 2015. Generation and design heuristics for zonal express services. Transp. Res. Part E Logist. Transp. Rev. 79, 201–212.

Leiva, C., Muñoz, J.C., Giesen, R., Larrain, H., 2010. Design of limited-stop services for an urban bus corridor with capacity constraints. Transp. Res. Part B Methodol. 44, 1186–1201.

Li, S.., Zheng, Y., Li, K., Wu, Y., Hedrick, J.., Gao, F., Zhang, H., 2017. Dynamical modeling and distributed control of connected and automated vehicles: Challenges and opportunities. IEEE Intell. Transp. Syst. Mag. 9, 46–58.

Lin, S., Tian, H., 2020. Short-Term Metro Passenger Flow Prediction Based on Random Forest and LSTM, in: 2020 IEEE 4th Information Technology, Networking, Electronic and Automation Control Conference (ITNEC). IEEE, pp. 2520–2526.

Liu, H., Xu, H., Yan, Y., Cai, Z., Sun, T., Li, W., 2020. Bus Arrival Time Prediction Based on LSTM and Spatial-Temporal Feature Vector. IEEE Access 8, 11917–11929.

Miller, P., De Barros, A.G., Kattan, L., Wirasinghe, S.C., 2016. Public Transportation and Sustainability: A Review. KSCE J. Civ. Eng. 20, 1076–1083.

Muñoz, J.C., Cortés, C.E., Giesen, R., Sáez, D., Delgado, F., Valencia, F., Cipriano, A., 2013. Comparison of dynamic control strategies for transit operations. Transp. Res. Part C Emerg. Technol. 28, 101–113.

Nassir, N., Hickman, • Mark, Ma, Z.-L., Hickman, M., 2015. Activity detection and transfer identification for public transit fare card data. Transportation (Amst). 42, 683–705.

Parsa, A.B., Chauhan, R.S., Taghipour, H., Derrible, S., Mohammadian, A., 2019. Applying Deep Learning to Detect Traffic Accidents in Real Time Using Spatiotemporal Sequential Data. arXiv Prepr. arXiv:1912.

Pezeshknejad, P., Monajem, S., Mozafari, H., 2020. Evaluating sustainability and land use integration of BRT stations via extended node place model, an application on BRT stations of Tehran. J. Transp. Geogr. 82, 102626.

Qi, J., Yang, L., Di, Z., Li, S., Yang, K., Gao, Y., 2018. Integrated optimization for train operation zone and stop plan with passenger distributions. Transp. Res. Part E Logist. Transp. Rev. 109, 151–173.

Shang, P., Li, R., Liu, Z., Yang, L., Wang, Y., 2018. Equity-oriented skip-stopping schedule optimization in an oversaturated urban rail transit network. Transp. Res. Part C Emerg. Technol. 89, 321–343.

Sharma, H., Jalal, A.S., 2020. Incorporating external knowledge for image captioning using CNN and LSTM. Mod. Phys. Lett. B 2050315.

Siami-Namini, S., Namin, A.S., 2018. Forecasting economics and financial time series: ARIMA vs. LSTM. arXiv Prepr. arXiv1803.06386.

Soltanpour, A., Mesbah, M., Habibian, M., 2020. Customer satisfaction in urban rail: a study on transferability of structural equation models. Public Transp. 12, 123–146.

Sun, A., Hickman, M., 2005. The real-time stop-skipping problem. J. Intell. Transp. Syst. 9, 91–109.

Vuchic, V.R. 1999, 1999. Transportation for Livable Cities, Center for Urban Policy Research, New Brunswick, NJ.

Wagenaar, J., Kroon, L., Fragkos, I., 2017. Rolling stock rescheduling in passenger railway transportation using dead-heading trips and adjusted passenger demand. Transp. Res. Part B Methodol. 101, 140–161.

Wang, Yihui, De Schutter, B., Van Den Boom, T.J.J., Ning, B., Tang, T., 2014. Efficient Bilevel approach for urban rail transit operation with stop-skipping. IEEE Trans. Intell. Transp. Syst. 15, 2658–2670.

Wang, Y., Schutter, B.D., Boom, T., Ning, B., Tang, T., 2014. Origin-Destination Dependent Train Scheduling Problem with Stop-Skipping for Urban Rail Transit Systems. undefined.

Y.Lu, B.Z. and, n.d. Research on dwelling time modeling of urban rail transit. Traffic Transp., vol. 27, no. 12, pp. 48–52, Dec. 2011.

Zhang, H., He, J., Bao, J., Hong, Q., Shi, X., 2020. A Hybrid Spatiotemporal Deep Learning Model for Short-Term Metro Passenger Flow Prediction. J. Adv. Transp. 2020.

Zhang, P., Sun, H., Qu, Y., Yin, H., Jin, J.G., Wu, J., 2021. Model and algorithm of coordinated flow controlling with station-based constraints in a metro system. Transp. Res. Part E Logist. Transp. Rev. 148, 102274.

Zhang, P., Sun, Z., Liu, X., 2017. Optimized Skip-Stop Metro Line Operation Using Smart Card Data.

Zolfaghari, S., Azizi, N., Jaber, M.Y., 2004. A model for holding strategy in public transit systems with real-time information. Int. J. Transp. Manag. 2, 99–110.